\documentclass[journal]{IEEEtran}
\usepackage{threeparttable}
\usepackage{xcolor,soul,framed} 
\colorlet{shadecolor}{yellow}
\usepackage[pdftex]{graphicx}
\graphicspath{{../pdf/}{../jpeg/}}
\DeclareGraphicsExtensions{.pdf,.jpeg,.png}
\usepackage{subcaption}
\usepackage{adjustbox}
\usepackage{booktabs}
\usepackage{array}
\usepackage{mdwmath}
\usepackage{mdwtab}
\usepackage{eqparbox}
\usepackage{url}
\usepackage{cite}
\usepackage{multirow}
\usepackage{commath}
\usepackage{longtable}
\usepackage{relsize}
\usepackage{amssymb}
\usepackage{amsmath, bm}
\usepackage{bbm}
\usepackage{amsfonts} 
\usepackage[switch]{lineno}
\usepackage{parskip}
\usepackage{rotating}
\usepackage{color}
\usepackage{booktabs,multirow,makecell,rotating,siunitx,adjustbox}
\usepackage{algorithm}
\usepackage{algpseudocode}
\newtheorem{definition}{Definition}
\usepackage{subcaption}
\begin{document}
\bstctlcite{IEEEexample:BSTcontrol}
    \title{Synergistic Fusion of Topological Structure and Temporal Semantics of Mobility for Urban
Region Embedding}
\author{
\IEEEauthorblockN{
Namwoo Kim\IEEEauthorrefmark{1},
Jeeyun Chang\IEEEauthorrefmark{2},
Kanghoon Lee\IEEEauthorrefmark{3},
and Yoonjin Yoon\IEEEauthorrefmark{1}\IEEEauthorrefmark{4}
\thanks{This work has been submitted to the IEEE for possible publication. Copyright may be transferred without notice, after which this version may no longer be accessible.}
}
\\[3pt]
\IEEEauthorblockA{
\IEEEauthorrefmark{1}Urban AI Institute, Korea Advanced Institute of Science and Technology (KAIST), Daejeon, Republic of Korea
\quad
\IEEEauthorrefmark{2}Graduate School of Data Science, KAIST, Daejeon, Republic of Korea
\\
\IEEEauthorrefmark{3}Department of Industrial Engineering and Systems, KAIST, Daejeon, Republic of Korea
\\
\IEEEauthorrefmark{4}Department of Civil and Environmental Engineering,
KAIST, Daejeon, Republic of Korea
}
}

\markboth{Preprint.}{}


\maketitle

\begin{abstract} 
Urban region embeddings have shown promising results in diverse urban sensing tasks such as crime, income, and service-call prediction. Recent methods improve representation quality by integrating mobility data with auxiliary modalities, using cross-view attention or contrastive objectives to align heterogeneous features into a unified region representation. However, leveraging the temporal dynamics of human mobility remains under-explored. Regional inflow and outflow fluctuate throughout the day, and inter-region connections emerge, persist, and dissolve over time. Moreover, prevailing fusion strategies combine views additively and miss the joint signal that emerges only when views co-occur. To address these gaps, we propose Mobility Stream–Structure Synergy (MoSS), which derives complementary views from mobility data: a Sequence view that preserves each region’s hourly inflow/outflow profile, and a Structure view based on zigzag persistence diagrams that capture how regional connectivity emerges, persists, and dissolves over time. A synergy module then extracts emergent representations from the co-occurrence of these views through multi-degree interactions, explicitly capturing higher-order signal across views. Extensive experiments on New York City and Chicago show that MoSS achieves state-of-the-art performance across three downstream tasks using mobility data alone, outperforming baselines that rely on auxiliary modalities. 
\end{abstract}


%


\section{Introduction}

\begin{figure}[htbp]
    \centering
\includegraphics[width=\columnwidth]{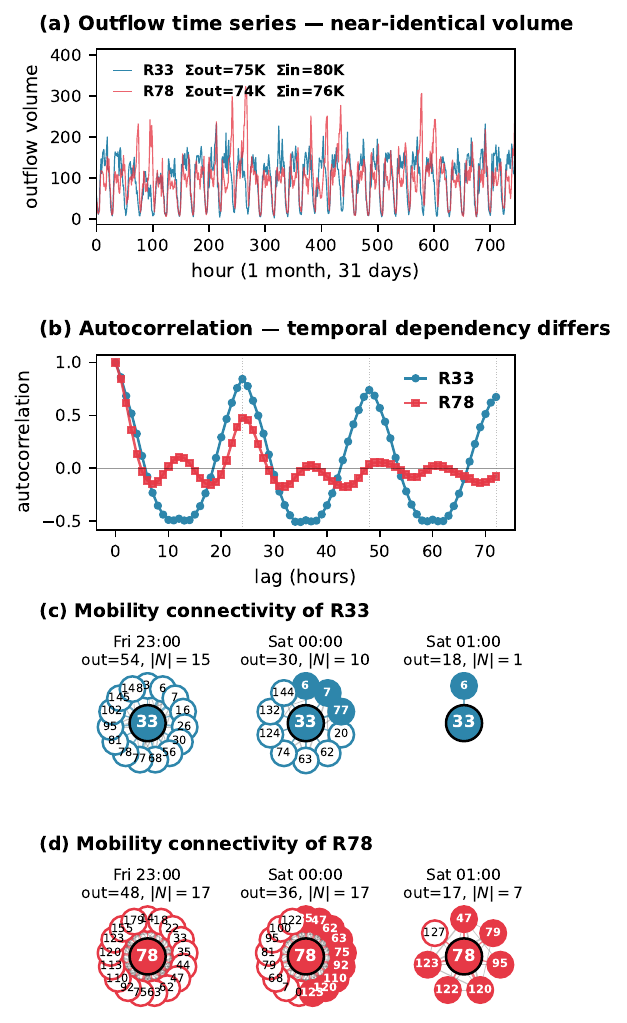}
    \caption{
        \textbf{Same volume, different connectivity dynamics.}
        Two Manhattan regions, $R_{33}$ and $R_{78}$, with near-identical
        aggregate mobility over the 31-day study period.
        \textbf{(a)}~Outflow time series over the full month: the two
        regions are comparable in aggregate volume.
        \textbf{(b)}~Autocorrelation up to a 72\,h lag: $R_{33}$
        exhibits a strong $24$\,h periodicity while $R_{78}$ shows a
        much weaker oscillation --- their temporal dependency
        structures differ markedly.
        \textbf{(c,\,d)}~Inter-region connectivity at three consecutive
        hours (Fri 23:00 $\rightarrow$ Sat 01:00) over which the two
        regions also share similar hourly outflow ($54/48 \to 30/36 \to 18/17$ trips).
        Despite this volume parity, neighborhood
        dynamics diverge sharply. Solid nodes denote neighbors persistent from the previous
        snapshot; outlined nodes are new entrants.
    }
    \label{fig:motivation_pair}
\end{figure}

\IEEEPARstart{U}{rban} region embedding maps each city region to a latent vector, enabling downstream tasks such as crime prediction, region popularity estimation, population density inference, land-use clustering, and socioeconomic analysis~\cite{survey, hdge, zemob, mvure, mgfn, hugat, recp, li2022predicting}. A region, however, cannot be characterized in isolation: its functional identity emerges from \emph{how} it interacts with other regions --- residential blocks outflow to business districts in the morning, commercial zones absorb inflows on weekends, and so on. To capture such interactions, human mobility data have been widely adopted, since every trip between two regions is an observable act of interaction that, in aggregate, encodes the functional organization of the city ~\cite{yuan2012discovering, zheng2014urban}.

A substantial body of work uses human mobility as the primary signal for region embedding. Early approaches modeled origin-destination (OD) co-occurrence with skip-gram-style objectives or random walks on flow graphs~\cite{hdge, zemob}. Subsequent work reconstructs OD matrices or conditional trip distributions~\cite{hugat, mvure, mgfn}, and the most recent approaches apply contrastive learning over inflow/outflow embeddings~\cite{remvc, recp}. These studies collectively establish that mobility carries rich functional information.

A complementary line of work enriches region representations by combining mobility with auxiliary sources --- POIs, check-ins, road networks --- and proposes a variety of fusion strategies. Prior studies have explored diverse strategies for multi-view urban representation learning. Early approaches focused on learning view-shared representations and aggregating them with adaptive weighting schemes~\cite{mvure}. Subsequent work modeled mobility patterns through graph clustering and cross-view attention mechanisms~\cite{mgfn}, while others built heterogeneous region graphs with prompt-based task adaptation~\cite{hrep}. More recent methods have emphasized fine-grained fusion and consistency learning, such as dual attention over regions and features~\cite{hafusion} and mutual-information-based consistency objectives across views~\cite{recp}. Recent advances further leverage graph pre-training, task-aware prompting~\cite{gurpp}, and structure-aware contrastive learning for joint multi-view representation learning~\cite{mvjc}. A concurrent line of work explicitly separates cross-view shared and view-private signals in this multi-modal setting~\cite{comsre}. Collectively, these methods show that the fusion of heterogeneous urban signals can improve downstream performance.

Despite these advances, current approaches leave two gaps under-explored. The first concerns how mobility itself is modeled. Mobility-oriented methods typically model mobility either as a static graph or as independently processed temporal snapshots, so they fail to capture both region-specific temporal dynamics---weekday/weekend periodicity, peak/off-peak transitions, and autocorrelation patterns---and the evolving connectivity structure among regions, where inter-region relations emerge, persist, and disappear over time~\cite{wang2018learning}. Figure~\ref{fig:motivation_pair} makes both facets concrete on a Manhattan pair: two regions with comparable aggregate outflow magnitudes nevertheless differ sharply in temporal dependency (Figure~\ref{fig:motivation_pair}(a, b)) as well as in inter-region connectivity (Figure~\ref{fig:motivation_pair}(c, d)).

The second gap concerns how views are fused. Even when both temporal and structural views are made available, prevailing multi-view fusion strategies---attention-based aggregation~\cite{mvure, mgfn, hafusion, hrep} or contrastive alignment~\cite{recp, mvjc}---combine views additively and tend to miss the co-occurrence-based signal that emerges only when the two are considered jointly. A region's functional identity often depends on the specific co-occurrence of its temporal rhythm and its connectivity neighborhood, so capturing this multiplicative interaction between temporal and structural patterns is essential for translating these two complementary streams into an effective region embedding that jointly reflects both signals.

To address these gaps, we propose Mobility Stream–Structure Synergy (MoSS), a region embedding framework that captures both per-region temporal dynamics and the evolving inter-region mobility structure from a single mobility stream, and explicitly models cross-view synergy.
MoSS consists of two complementary streams. A Sequence stream captures each region's \emph{temporal semantics} by modeling its hourly inflow and outflow series with dilated convolutions that span daily and weekly periodicities.
A Structure stream captures each region's \emph{topological structure} by tracking the evolution of its connectivity neighborhood through \emph{zigzag persistence diagrams}~\cite{zigzag}.
The two streams produce complementary views per region, which we fuse via a synergy module built on a shared--private feature decomposition together with multi-degree interactions. Empirical evaluation on New York City and Chicago shows that MoSS achieves state-of-the-art region embedding performance from a single mobility input, without relying on any auxiliary modality.

To summarize, our contributions are as follows.
\begin{itemize}
    \item We propose a dual-stream mobility representation consisting of a sequence stream and a structure stream --- to the best of our knowledge, the first application of zigzag persistent homology to urban region embedding.
    \item To model higher-order cross-view interactions, we introduce a synergy module that combines shared--private decomposition with multi-degree multiplicative interactions.
    \item Extensive experiments on New York City and Chicago show that \textbf{MoSS} achieves state-of-the-art urban region embedding using mobility data alone, outperforming baselines that rely on auxiliary modalities such as POIs, check-ins, or land use.
\end{itemize}

\section{Related Work}
\label{sec:related}
\subsection{Human Mobility in Urban Region Embedding}

Human mobility has long been a core signal for urban region embedding. Early studies such as ZE-Mob~\cite{zemob} and HDGE~\cite{hdge} primarily modeled mobility using static origin--destination (OD) statistics or transition graphs. MGFN~\cite{mgfn} extended this line of work by introducing multiple mobility patterns derived from grouped OD snapshots, but these patterns were still constructed from static aggregations and did not capture the continuous temporal dynamics of mobility flows. Subsequent studies incorporated additional urban modalities such as POIs, land use, satellite imagery, and street views \cite{read, knowcl,remvc,mvure,hrep,hgaurban,hugat,hafusion, zhang2019unifying, li2023urban}. Other recent approaches explored prompt learning and contrastive learning frameworks for urban representation learning \cite{hrep,gurpp,flexireg,recp,mvjc}.

In summary, mobility is still often modeled as a static signal, typically using aggregated OD matrices, transition graphs, or temporally grouped snapshots. While these representations have proven effective for capturing large-scale mobility structure, they generally summarize observations over time and therefore provide only limited access to the temporal continuity and dependency inherent in human movement patterns. As a result, the dynamic evolution of inter-region interactions, including how connections form, persist, and change across hours and days, remains less explored. To address this gap, we model mobility as fine-grained temporal dynamics and capture both its flow-level temporal patterns and its evolving connectivity topology within a unified framework.

\subsection{Multi-View Approaches in Urban Region Embedding}

A growing body of work leverages diverse modalities and views to learn comprehensive and semantically rich urban region embeddings~\cite{Fu_Wang_Du_Wu_Li_2019}. These multi-view methods cluster into three fusion families. Attention-based fusion~\cite{mvure, mgfn, hafusion, hrep} combines per-view embeddings through input-dependent weighted sums, often arranged hierarchically across intra-view, inter-view, and region levels. Contrastive alignment~\cite{recp, mvjc, gurpp, remvc} reshapes view-specific embeddings via cross-view agreement objectives---mutual information maximization, structure-aware contrastive losses, or prompt-based self-supervision---to enforce consistency. Shared--private decomposition, established for domain adaptation and multimodal sentiment analysis~\cite{domainseparation, misa}, has been recently adapted to urban region embedding by ComSRE~\cite{comsre} on mobility and POI views, instantiated with a contrastive alignment and a differential orthogonality penalty.

These three families, however, remain limited in modeling higher-order cross-view interactions. Informative patterns that emerge only through the joint presence of multiple views are not explicitly represented in additive fusion~\cite{ZHAO201743}. Prior work has shown that explicitly modeling multiplicative interactions can improve multimodal representation learning~\cite{zadeh-etal-2017-tensor, Yu_2017_ICCV, liu-etal-2018-efficient-low}. Building on this intuition, we introduce a \emph{synergy module} that models cross-view interactions, including pairwise (second-order) and triple-wise (third-order) multiplicative terms across views. Unlike attention or contrastive alignment, synergy module explicitly incorporates these higher-order interaction terms into the representation pathway, providing a compositional inductive bias for multi-view urban region embedding. 



\subsection{Topological Data Analysis for Time-Varying Data} 

Persistent homology, a central tool in topological data analysis (TDA), summarizes the multi-scale shape of data as a persistence diagram, and has been used as a complementary signal in deep representation learning, both for time series~\cite{topocl} and for tracking structural change in time-varying graphs~\cite{8365984}. However, a standard filtration grows monotonically, so it cannot represent features that vanish and re-emerge. Zigzag persistence~\cite{zigzag} removes this restriction by allowing inclusions in alternating directions, which allows such features to be tracked directly (Sec.~\ref{sec:prelim-zigzag}), and a growing line of work applies it to temporal and graph-structured data. It has been used to detect bifurcations in dynamical systems~\cite{tymochko2020using} and to summarize temporal networks as a single zigzag diagram, recovering daily and weekly commuting periodicities in transportation that connectivity and centrality statistics miss~\cite{myers2023temporal}. More recent work integrates zigzag persistence into graph neural networks, either as a time-aware topological layer~\cite{pmlr-v139-chen21o} or as a compact filtration-curve summary used for time-series forecasting~\cite{chen2022time}. 

Across these settings, zigzag persistence has proven effective in capturing how connectivity evolves over time, but it has not yet been used for urban region embedding.

\section{Preliminaries}
\label{sec:prelim}

This section introduces the topological tools that MoSS uses to summarize the temporal evolution of inter-regional mobility. We refer to \cite{edelsbrunner2010computational}, \cite{zigzag}, and \cite{myers2023temporal} for a complete treatment.

\subsection{Clique Complex of a Graph}
\label{sec:prelim-complex}

A \emph{simplicial complex} $\mathcal{K}$ over a vertex set $\mathcal{V}$ is a collection of subsets (simplices) closed under taking subsets. A subset of size $k{+}1$ is a $k$-simplex, so that vertices, edges, and triangles are the $0$-, $1$-, $2$-, and higher-dimensional simplices respectively.

\begin{definition}[Clique complex]
\label{def:clique}
Let $G = (\mathcal{V}, \mathcal{E})$ be a finite undirected graph. The \emph{clique complex} of $G$ up to dimension $D$ is
\begin{equation}
    \mathcal{X}_D(G)
    \;=\; \bigl\{ \sigma \subseteq \mathcal{V} \,:\, 1 \leq |\sigma| \leq D{+}1, \;\sigma \text{ is a clique in } G \bigr\}.
    \label{eq:clique-complex}
\end{equation}
\end{definition}

\subsection{Homology and Betti Numbers}
\label{sec:prelim-homology}

For each $k \geq 0$, let $C_k(\mathcal{K})$ be the $\mathbb{F}_2$-vector space spanned by the $k$-simplices of $\mathcal{K}$. The boundary map $\partial_k\colon C_k(\mathcal{K}) \to C_{k-1}(\mathcal{K})$ sends a $k$-simplex to the formal sum of its $(k{-}1)$-faces, and satisfies $\partial_{k-1} \circ \partial_k = 0$. Thus, the $k$-th homology group is
\begin{equation}
    H_k(\mathcal{K}) \;=\; \ker(\partial_k) \,\big/\, \mathrm{im}(\partial_{k+1}),
    \label{eq:homology}
\end{equation}
i.e., the cycles modulo the boundaries.

\begin{definition}[Betti number]
\label{def:betti}
The $k$-th \emph{Betti number} of a simplicial complex $\mathcal{K}$ is the dimension of its $k$-th homology group over $\mathbb{F}_2$,
\begin{equation}
    \beta_k(\mathcal{K}) \;=\; \dim H_k(\mathcal{K}).
    \label{eq:betti}
\end{equation}
Geometrically, $\beta_k$ counts the number of independent $k$-dimensional holes in $\mathcal{K}$: $\beta_0$ is the number of connected components (separate pieces), $\beta_1$ the number of $1$-dimensional loops (tunnels through the complex), and $\beta_2$ the number of enclosed $2$-dimensional cavities.

\end{definition}

A static $\beta_k$ is one snapshot of structure. Capturing how features appear and disappear over time requires a notion of \emph{persistence}.

\subsection{Persistent Homology}
\label{sec:prelim-ph}

A \emph{filtration} is a nested sequence
\begin{equation}
    \mathcal{K}^{(1)} \subseteq \mathcal{K}^{(2)} \subseteq \cdots \subseteq \mathcal{K}^{(T)}.
    \label{eq:filtration}
\end{equation}
Persistent homology tracks each topological feature as it appears and disappears across this filtration sequence, recording it as a point $(b,d)$, where $b$ denotes the filtration index at which the feature appears and $d$ the index at which it disappears. The collection of such points is the \emph{persistence diagram} in dimension $k$. 

A filtration only ever grows, so it cannot represent features that disappear and reappear. This is a structural property of urban mobility where edges between regions emerge during peak hours and dissolve afterwards~\cite{myers2023temporal, 8365984}.

\subsection{Zigzag Persistent Homology}
\label{sec:prelim-zigzag}

Zigzag persistent homology~\cite{zigzag} replaces the one-way filtration of Eq.~\eqref{eq:filtration} by a sequence of inclusions in alternating directions,
\begin{equation}
    \mathcal{K}^{(1)}
    \hookrightarrow \mathcal{K}^{(1{,}2)}
    \hookleftarrow  \mathcal{K}^{(2)}
    \hookrightarrow \cdots
    \hookleftarrow  \mathcal{K}^{(T)},
    \label{eq:zigzag-skeleton}
\end{equation}
where we take $\mathcal{K}^{(t,t{+}1)} = \mathcal{K}^{(t)} \cup \mathcal{K}^{(t{+}1)}$. The forward inclusion $\mathcal{K}^{(t)} \hookrightarrow \mathcal{K}^{(t,t{+}1)}$ adds simplices that appear in $\mathcal{K}^{(t{+}1)}$ but not in $\mathcal{K}^{(t)}$, and the backward inclusion $\mathcal{K}^{(t{+}1)} \hookrightarrow \mathcal{K}^{(t,t{+}1)}$ adds simplices that exist in $\mathcal{K}^{(t)}$ but not in $\mathcal{K}^{(t{+}1)}$. The output is again a multiset of points $(b, d)$, but indexed along the zigzag positions rather than along a monotone parameter, so that features which vanish and re-emerge are represented natively rather than lost.

Two structural properties of urban mobility motivate the use of zigzag persistence. First, edges in inter-regional mobility are inherently \emph{transient}, so the standard filtration of Eq.~\eqref{eq:filtration} cannot represent dissolution. Second, the macro-level role of a region depends on \emph{when} its connections appear and dissolve, rather than only on the static set of connections that eventually exist.

\section{Method}
\label{sec:method}

\subsection{Model Overview}
\label{sec:overview}

\begin{figure*}[htbp]
    \centering
    \includegraphics[width=\linewidth]{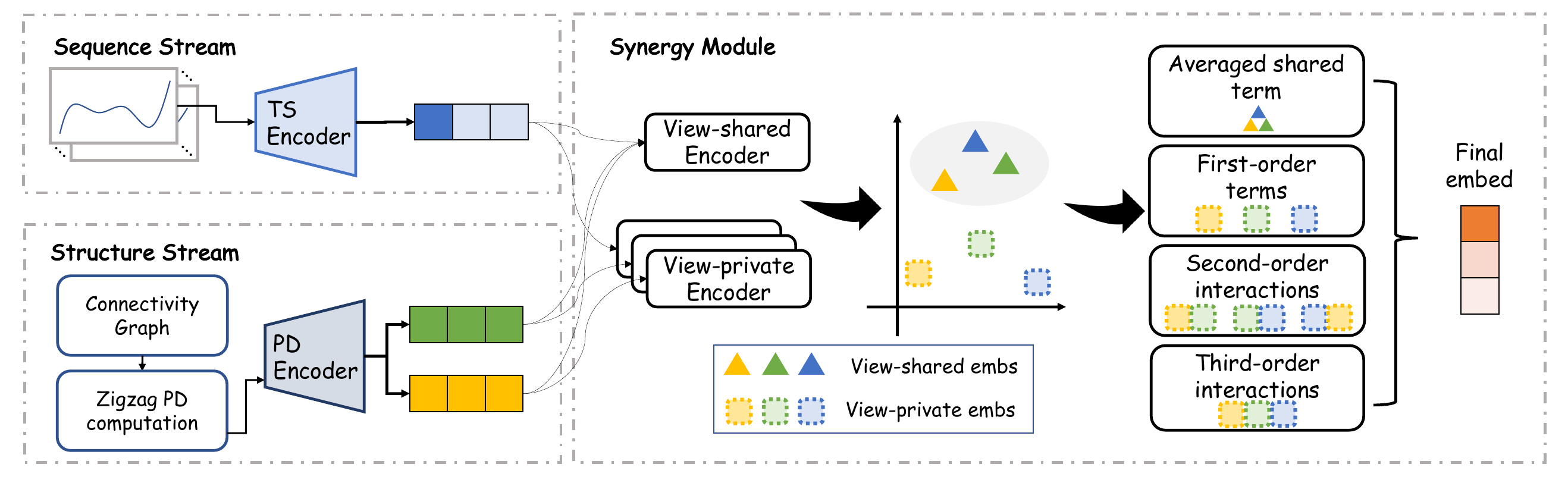}
    \caption{
        \textbf{Overview of MoSS.}
        \textit{Left.} Two complementary streams encode each region's mobility. The \textbf{Sequence (TS) view} (top) feeds the per-region hourly inflow/outflow series into a temporal convolutional
        encoder (TS Encoder), and the \textbf{Structure (PD) view} (bottom) constructs the region-centric connectivity-graph at each time step, computes its zigzag persistence diagram, and encodes the
        diagram with a permutation-invariant set encoder (PD Encoder). The structure encoder produces source- and destination-side view embeddings.
        \textit{Right.} The \textbf{synergy module} fuses the three views. It first applies a \emph{shared--private decomposition} that factors each view into a cross-view shared embedding (filled triangles) and a view-private embedding (dashed circles), and then aggregates the resulting embeddings through \emph{multi-degree interaction} that captures first-order, second-order, and third-order interactions in a single per-region embedding.
    }
    \label{fig:model_overview}
\end{figure*}

\textsc{MoSS} represents each region from two complementary viewpoints of the same mobility stream and fuses them through a synergy module (Figure~\ref{fig:model_overview}). The
\emph{Sequence stream} (Sec.~\ref{sec:ts-encoder}) encodes each region's hourly inflow/outflow series with a dilated TCN. The \emph{Structure stream} (Sec.~\ref{sec:connectivity-graph}--\ref{sec:pd-encoder}) encodes how the inter-region connections around a region emerge, persist, and dissolve over the observation window using zigzag persistence. The resulting view tensors are then composed by the \emph{synergy module} (Sec.~\ref{sec:synergy}), which internally factors each view into shared and view-private embeddings and aggregates them through multi-degree interaction that captures first-order, second-order, and third-order interactions across views.

\subsection{Sequence Stream}
\label{sec:ts-encoder}

Each region $i$ is associated with two observed series of trip volumes (inflow and outflow), recording how many trips leave and arrive at the region over the full observation window of length $T$. We stack the two normalized series along the channel axis to form a two-channel input $\mathbf{x}_i = [\bar{\mathbf{x}}_i^{\mathrm{inflow}};\, \bar{\mathbf{x}}_i^{\mathrm{outflow}}] \in \mathbb{R}^{2 \times T}$.

A single dilated temporal convolutional network~\cite{vandenoord2016wavenet, bai2018empirical} processes both flow directions jointly. After a $1{\times}1$ projection of the two-channel input to hidden width $C_h$, the series passes through a stack of $L_{\mathrm{tcn}}$ residual blocks; the $\ell$-th block applies two width-preserving 1D convolutions of kernel size $k$ with dilation $2^{\ell}$, with a GELU pre-activation before each convolution and a residual shortcut around the pair,
\begin{equation}
    \mathbf{h}_\ell
    = \mathbf{h}_{\ell-1}
        + \mathrm{Conv}^{(2)}_{2^\ell}\!\Bigl(
            \mathrm{GELU}\bigl(
                \mathrm{Conv}^{(1)}_{2^\ell}\!\bigl(\mathrm{GELU}(\mathbf{h}_{\ell-1})\bigr)
            \bigr)
          \Bigr).
    \label{eq:tcn-block}
\end{equation}
Exponentially growing dilation gives the topmost block a receptive field that spans both daily and weekly periodicities using only a logarithmic number of layers, while the residual path keeps gradients well behaved. A final $1{\times}1$ projection maps the channel dimension to $D$-dimension, after which we reduce the time axis by max-pooling to produce a sequence-view embedding
\begin{equation}
    \mathbf{V}^{\mathrm{seq}}_i
    = \mathrm{maxpool}_t\!\Bigl(
        \mathrm{Conv}^{(\mathrm{out})}\bigl(\mathbf{h}_{L_{\mathrm{tcn}}}\bigr)
      \Bigr) \in \mathbb{R}^{D}.
    \label{eq:ts-embed}
\end{equation}

\subsection{Structure Stream: Connectivity-graph Construction}
\label{sec:connectivity-graph}

\begin{figure*}[htbp]
    \centering
    \resizebox{0.8\linewidth}{!}{%
        \includegraphics{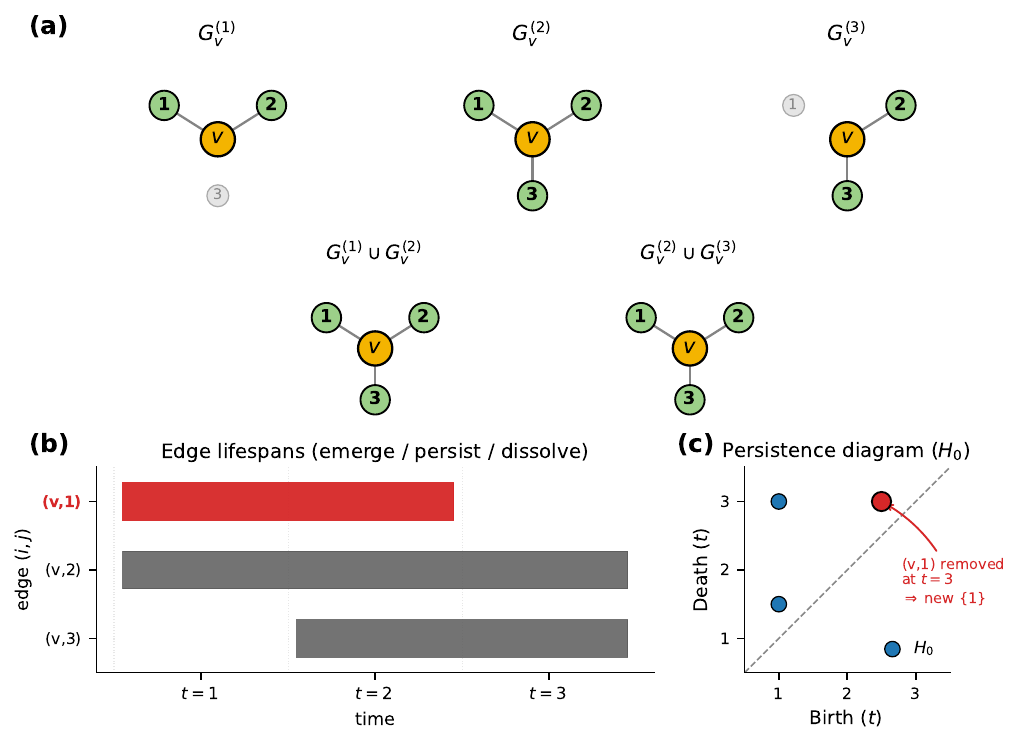}%
    }
    \caption{
        \textbf{Structure stream: from a temporal sequence of connectivity-graphs
        to a persistence diagram}.
        \textbf{(a)} Connectivity-graph $G_v^{(t)}$ at three consecutive time indices (top row), and the union graph $G_v^{(t)} \cup G_v^{(t+1)}$ between every consecutive pair (middle row); filled green nodes are neighbors active at that time, faded gray nodes are inactive.
        \textbf{(b)} Per-edge lifespans across the window. The bar segments mark the time indices at which the edge is present, exposing the emerge / persist / dissolve dynamics that no single snapshot captures.
        \textbf{(c)} The resulting zigzag persistence diagram with $H_0$ features; the diagonal marks zero lifespan.
    }
    \label{fig:structure_stream}
\end{figure*}

Figure~\ref{fig:structure_stream} gives a visual roadmap of the structure stream: a temporal sequence of region-centric connectivity-graphs (Figure~\ref{fig:structure_stream}a) is summarized as per-edge lifespans (Figure~\ref{fig:structure_stream}b), and zigzag persistent homology turns those lifespans into a persistence diagram (Figure~\ref{fig:structure_stream}c) that the structure stream then encodes.

Starting from the raw OD matrices $\{\mathbf{F}^{(t)}\}_{t=1}^{T}$ with $\mathbf{F}^{(t)} \in \mathbb{R}^{N \times N}$, we condense the stream into a shorter representative period of length $T'$ (e.g., a typical week of hourly snapshots) by averaging OD matrices that share the same within-period phase across repeated cycles. The resulting averaged matrices are then binarized so that each entry records only whether a flow exists between two regions.

For each region $i$, the connectivity-graph $G_i^{(t)}$ at time index $t$ is the subgraph of $G^{(t)}$ induced by $i$ and the regions connected to it (Figure~\ref{fig:structure_stream}a). We construct two directional variants: the outflow graph $G_i^{\mathrm{out},(t)}$, which includes edges where $i$ is the source, and the inflow graph $G_i^{\mathrm{in},(t)}$, which includes edges where region $i$ is the destination. These two graphs capture complementary aspects of the functional role of region $i$. The temporally aligned sequences $G_i^{\mathrm{in},(t)}$ and $G_i^{\mathrm{out},(t)}$ are then used as input to the zigzag persistence computation. For simplicity, we use $\circ \in \{\mathrm{out}, \mathrm{in}\}$ to denote either direction in what follows.

\subsection{Structure Stream: Zigzag Persistence Computation}
\label{sec:zigzag}

For each region $i$, we build a clique complex $\mathcal{K}_i^{\circ,(t)}$ over the connectivity-graph $G_i^{\circ,(t)}$ at every time index, and connect consecutive complexes through the zigzag diagram in Eq.~\eqref{eq:zigzag-skeleton}. Zigzag persistent homology in dimension 0 returns a persistence diagram $\mathrm{PD}_i^{\circ}$ that summarizes how the corresponding feature class (connected components) of the mobility neighborhood of $i$ emerges, persists, and dissolves over the time window. 

Figure ~\ref{fig:structure_stream} illustrates this construction on a four-node toy.
Figure ~\ref{fig:structure_stream} (a) shows the per-time graphs $G_v^{(t)}$ for $t=1,2,3$, together with the consecutive unions $G_v^{(t)} \cup G_v^{(t+1)}$. These union graphs occupy the half-integer slots $t=1.5, 2.5$ along the filtration axis, sitting between the per-time snapshots so that edges added or removed across consecutive times are accounted for.
Figure ~\ref{fig:structure_stream} (b) renders each edge's lifespan as a horizontal bar. As edges enter or leave between consecutive complexes, the number of connected components ($\beta_0$, the rank of $H_0$) of the current $\mathcal{K}$ rises and falls, and the persistence diagram in (c) collects these changes into one point per component: birth $b$ marks the step where a new component first separates out, death $d$ marks the step where it merges back into an older one (or is pinned to $d=T$ if it survives the window).

The red point $(b,d)=(2.5, 3)$ traces a single such event. Edge $(v,1)$ is present through $K_2$ and the union $K_{2,3}$ but disappears at $K_3$, which disconnects node $1$ from $v$'s component. The resulting singleton is born at the preceding union step $b=2.5$ and, since $(v,1)$ does not reappear, persists to the window's end $d=T=3$.

The other two points are read similarly. $(1, 3)$ is $v$'s own component, born at $K_1$ and alive throughout the window, while $(1, 1.5)$ is the briefly-isolated node $3$ that merges into $v$'s component as soon as edge $(v,3)$ first enters at $K_{1,2}$.


We restrict the computation to $H_0$ because the structure stream targets \emph{connectivity}. $\beta_0$ counts the connected components of a region's mobility neighborhood, so it responds directly to whether regions remain linked to its neighbors. When a region becomes disconnected from its neighborhood, a new component appears, and when regions reconnect, components merge. Therefore, tracking $H_0$ across the zigzag captures the connectivity changes that motivate the structure stream, while higher-dimensional features ($\beta_1$ and above) are not needed for this connectivity-focused purpose.

\subsection{Structure Stream: Persistence-diagram Encoder}
\label{sec:pd-encoder}

The zigzag persistence computation returns, for each region $i$ and direction $\circ \in \{\mathrm{out}, \mathrm{in}\}$, a multiset of birth--death pairs $(b^{\circ}, d^{\circ})$ that characterize the emergence and disappearance of connected components over time in the mobility neighborhood of region $i$ along direction $\circ$. We augment each pair with its lifespan $l^{\circ} = d^{\circ} - b^{\circ}$, yielding the per-region, per-direction persistence multiset
\begin{equation}
\bigl\{p_{i,1}^{\circ}, \ldots, p_{i,M_i}^{\circ}\bigr\},
    \qquad
    p_{i,j}^{\circ} = (b_{i,j}^{\circ},\, d_{i,j}^{\circ},\, l_{i,j}^{\circ}) \in \mathbb{R}^3,
    \label{eq:pd-multiset}
\end{equation}
where $M_i$ denotes the cardinality. Our goal is to map this multiset to a fixed-size vector $\mathbf{V}^{\circ}_i \in \mathbb{R}^D$.

\paragraph{Permutation invariance.}
Points in the persistence diagram carry no canonical ordering, since the birth--death pairs index topological features that all coexist during the zigzag and have no natural total order. An order-sensitive encoder would treat the diagram as an order sequence rather than a set, so the resulting embedding would depend on an arbitrary indexing convention rather than on the topological information the diagram actually encodes. To rule this out by construction, the encoder $f^{\circ}_{\theta}$ must satisfy, for any multiset $\{p_1, \ldots, p_M\}$ of diagram points,


\begin{equation}
    f^{\circ}_{\theta}\!\bigl(\{p_{\pi(1)}, \ldots, p_{\pi(M)}\}\bigr)
    \;=\; f^{\circ}_{\theta}\!\bigl(\{p_{1}, \ldots, p_{M}\}\bigr)
    \qquad \forall\, \pi \in S_{M}.
    \label{eq:perm-invariance}
\end{equation}

\paragraph{Encoder.}
The canonical way to build a function that satisfies Eq.~\eqref{eq:perm-invariance} is to compose a per-point transformation shared across all points with a symmetric aggregator over the resulting features~\cite{deepset, pointnet}. We instantiate this as
\begin{equation}
    \mathbf{V}^{\circ}_i
    \;=\; f^{\circ}_{\theta}\!\bigl(\{p_{i,1}^{\circ}, \ldots, p_{i,M_i}^{\circ}\}\bigr)
    \;=\; \zeta\!\Bigl(\bigl\{\phi(p_{i,j}^{\circ})\bigr\}_{j=1}^{M_i}\Bigr)
    \;\in\; \mathbb{R}^{D},
    \label{eq:topo-encoder}
\end{equation}
where $\phi_\theta\colon \mathbb{R}^3 \to \mathbb{R}^D$ is a stack of $1\!\times\!1$ convolutions along the point axis with ReLU activations --- equivalent to applying the same MLP to every diagram point independently --- and $\zeta$ is coordinate-wise max-pooling across the $M_i$ feature vectors. Both $\phi$ and $\zeta$ depend only on the multiset of their inputs and not on the order in which the points are presented, so $f_\theta$ satisfies Eq.~\eqref{eq:perm-invariance} by construction.

Through $f_\theta^{\circ}$ for $\circ \in \{\mathrm{out}, \mathrm{in}\}$, we obtain two embeddings $\mathbf{V}^{\mathrm{out}}$ and $\mathbf{V}^{\mathrm{in}}$ in $\mathbb{R}^{N \times D}$. Each row encodes when the connected components of a region's neighborhood emerge, persist, and disappear.

\subsection{Synergy Module}
\label{sec:synergy}
The two encoder streams produce three view embeddings per region: a sequence view $\mathbf{V}^{\mathrm{seq}}$ from the temporal series and two directional views $\mathbf{V}^{\mathrm{out}}, \mathbf{V}^{\mathrm{in}}$ from the zigzag persistence diagrams, collected as $\mathcal{V} = \{\mathrm{seq}, \mathrm{out}, \mathrm{in}\}$. We take \emph{synergy} to mean the useful features that arise only when two or more of these views co-occur --- signal that is uninformative, or not present at all, in any single view observed in isolation. Recovering such features requires that the fusion stage receive, from each view, only what that view contributes distinctively: if shared content were left inside each view, every cross-view product would repeatedly recombine the same redundant signal, causing the fused representation to be dominated by redundant cross-view correlations rather than genuinely complementary interactions. 

The synergy module therefore fuses the three views in two stages. First, a \emph{shared--private decomposition} splits each view into a cross-view shared embedding and a view-private embedding. Second, a \emph{multi-degree interaction} applies a low-rank tensor product over the three private embeddings, enabling the fused representation to capture higher-order interactions across views.

\paragraph{Shared--private decomposition.}
\label{sec:ss-decomp}
For each view embedding $\mathbf{V}^{v} \in \mathbb{R}^{D}$, a shared projection $g_{\mathrm{shared}}: \mathbb{R}^{D} \to \mathbb{R}^{D}$ with parameters tied across views and a view-private projection $g_{\mathrm{private}}^{v}: \mathbb{R}^{D} \to \mathbb{R}^{D}$ produce~\cite{domainseparation, misa}
\begin{equation}
\begin{aligned}
    \mathbf{S}^{v} &= g_{\mathrm{shared}}\bigl(\mathbf{V}^{v}\bigr) \in \mathbb{R}^{D}, \\
    \mathbf{P}^{v} &= g_{\mathrm{private}}^{v}\bigl(\mathbf{V}^{v}\bigr) \in \mathbb{R}^{D},
\end{aligned}
\qquad v \in \mathcal{V},
    \label{eq:ss-decomp}
\end{equation}
each implemented as a two-layer MLP (Linear--GELU--Dropout--Linear). Two auxiliary losses shape the decomposition. A cross-view Central Moment Discrepancy (CMD)~\cite{cmd} pulls the shared embeddings toward a common distribution by matching their first $K$ central moments across view pairs,
\begin{equation}
    \mathcal{L}_{\mathrm{align}}
        = \!\!\sum_{\{v,w\} \subset \mathcal{V}}\!\!
            \mathrm{CMD}_K\bigl(\mathbf{S}^{v},\, \mathbf{S}^{w}\bigr),
    \label{eq:align-loss}
\end{equation}
while an orthogonality penalty decorrelates $\mathbf{S}^{v} \perp \mathbf{P}^{v}$ within each view and $\mathbf{P}^{v} \perp \mathbf{P}^{w}$ across views,
\begin{equation}
    \mathcal{L}_{\mathrm{orth}}
        = \sum_{v \in \mathcal{V}} \mathrm{Decorr}\bigl(\mathbf{S}^{v}, \mathbf{P}^{v}\bigr)
        + \!\!\sum_{\{v,w\} \subset \mathcal{V}}\!\!
          \mathrm{Decorr}\bigl(\mathbf{P}^{v}, \mathbf{P}^{w}\bigr),
    \label{eq:orth-loss}
\end{equation}
where $\mathrm{Decorr}(\mathbf{A}, \mathbf{B}) = \|\tilde{\mathbf{A}}^{\!\top}\tilde{\mathbf{B}}\|_F^{2}$ is the squared Frobenius cross-product on row-centered, row-normalized matrices $\tilde{\mathbf{A}}, \tilde{\mathbf{B}}$. Together these objectives drive the decomposition toward its intended structure: $\mathcal{L}_{\mathrm{align}}$ collapses the three shared embeddings $\{\mathbf{S}^{v}\}_{v \in \mathcal{V}}$ toward a common cross-view representation that captures content present across all views, while $\mathcal{L}_{\mathrm{orth}}$ keeps each private $\mathbf{P}^{v}$ distinguishable from both its within-view shared embeddings and the other views' privates, so that $\mathbf{P}^{v}$ retains only signal specific to view $v$.

\paragraph{Multi-degree Interaction.}
\label{sec:syn-pool}
With the shared--private decomposition in place, the synergy module fuses only the three private embeddings $\{\mathbf p^{\mathrm{seq}},\, \mathbf p^{\mathrm{out}},\, \mathbf p^{\mathrm{in}}\}$, each in $\mathbb R^{D}$. Synergy, as we defined it above, requires interactions across views; to capture all such interactions of every degree within a single object, we first augment each private with a constant unit,
\begin{equation}
    \widetilde{\mathbf p}^{\,v}
    \;=\;
    [\,1\,;\,\mathbf p^{v}\,]
    \;\in\;
    \mathbb R^{D+1},
    \qquad v \in \mathcal V.
    \label{eq:aug-private}
\end{equation}

The outer product of the augmented privates yields the fusion tensor
\begin{equation}
    \mathcal T
    \;=\;
    \widetilde{\mathbf p}^{\,\mathrm{seq}}
    \otimes
    \widetilde{\mathbf p}^{\,\mathrm{out}}
    \otimes
    \widetilde{\mathbf p}^{\,\mathrm{in}}
    \;\in\;
    \mathbb R^{(D+1)\times(D+1)\times(D+1)}.
    \label{eq:fusion-tensor}
\end{equation}
Because the zeroth coordinate corresponds to the prepended constant unit, the tensor simultaneously encodes first-order terms (e.g., $p_i^{\mathrm{seq}}$), second-order interactions (e.g., $p_i^{\mathrm{seq}} p_j^{\mathrm{out}}$), and full third-order interactions $p_i^{\mathrm{seq}} p_j^{\mathrm{out}} p_k^{\mathrm{in}}$.

A synergy embedding is computed via a learned tensor projection, where a weight tensor $\mathcal W \in \mathbb R^{D \times (D+1)^{|\mathcal V|}}$ linearly combines the entries of $\mathcal T$ as follows:
\begin{equation}
    Y_{\mathrm{syn},\,m}
    \;=\;
    \sum_{i,\,j,\,k}
    \mathcal W_{m,\,i,\,j,\,k}\;
    \mathcal T_{ijk}.
    \label{eq:fusion-contraction}
\end{equation}
Direct parameterization is computationally expensive because $\mathcal W$ alone contains $D(D+1)^{|\mathcal V|}$ parameters. To obtain a more parameter-efficient formulation, we approximate $\mathcal W$ using a low-rank tensor factorization following~\cite{liu-etal-2018-efficient-low}:
\begin{equation}
\mathcal W
\approx
\sum_{r=1}^{R}
\mathbf w^{\mathrm{y}}_{r}
\otimes
\mathbf u^{\mathrm{seq}}_{r}
\otimes
\mathbf u^{\mathrm{out}}_{r}
\otimes
\mathbf u^{\mathrm{in}}_{r},
\label{eq:cp}
\end{equation}
collecting the view factors $\mathbf u^{v}_{r} \in \mathbb R^{D+1}$ into $\mathbf U^{v} \in \mathbb R^{R \times (D+1)}$ and the output factors $\mathbf w^{\mathrm{y}}_{r} \in \mathbb R^{D}$ into $\mathbf W_{\mathrm{y}} \in \mathbb R^{D \times R}$. Substituting~\eqref{eq:cp} into~\eqref{eq:fusion-contraction} yields an equivalent factorized form in which the fusion reduces to an element-wise Hadamard product across views in rank space,
\begin{equation}
    \mathbf z
    \;=\;
    \bigodot_{v \in \mathcal V}\,
    \mathbf U^{v}\,\widetilde{\mathbf p}^{\,v}
    \;\in\;
    \mathbb R^{R},
    \qquad
    \mathbf Y_{\mathrm{syn}}
    \;=\;
    \mathbf W_{\mathrm{y}}\,\mathbf z
    \;\in\;
    \mathbb R^{D}.
    \label{eq:syn-lmf}
\end{equation}
This compression preserves the multi-degree interaction structure described above. 

Finally, the synergy output $\mathbf Y_{\mathrm{syn}}$ is recombined with the shared content as follows:

\begin{equation}
    \mathbf{Y}
        = \mathrm{LayerNorm}\bigl(\mathbf{Y}_{\mathrm{syn}}\bigr)
        + \mathrm{LayerNorm}\bigl(\bar{\mathbf{S}}\bigr) \in \mathbb{R}^{D},
    \label{eq:fusion}
\end{equation}
where $\bar{\mathbf S}=\frac{1}{|\mathcal V|} \sum_{v \in \mathcal V}\mathbf S^{v}$ denotes the mean shared embeddings across views.

\subsection{Trip-distribution Loss}
\label{sec:trip-loss}

We supervise the learned embeddings using the empirical OD distribution $\mathbf M \in \mathbb R^{N\times N}$, obtained from the observed OD matrix.

The synergy embedding $\mathbf Y \in \mathbb{R}^{D}$ is mapped to a region representation $\mathbf Z = f_{\mathrm{head}}(\mathbf Y) \in \mathbb R^{N\times H}$ by a two-layer MLP head that expands the view dimension $D$ to the final embedding dimension $H$, and then projected into source and destination embeddings,
\begin{equation}
    \mathbf Z^{\mathrm{src}}
    =
    \mathbf Z\,\mathbf W_{\mathrm{src}}^{\!\top},
    \qquad
    \mathbf Z^{\mathrm{dst}}
    =
    \mathbf Z\,\mathbf W_{\mathrm{dst}}^{\!\top}.
    \label{eq:trip-heads}
\end{equation}
Then, the mobility distributions are computed as
\begin{equation}
    \widehat{\mathbf Q}^{\mathrm{out}}_{i,j}
    =
    \frac{
        \exp\!\bigl(
            \mathbf z_i^{\mathrm{src}}
            \!\cdot\!
            \mathbf z_j^{\mathrm{dst}}
        \bigr)
    }{
        \sum_{k=1}^{N}
        \exp\!\bigl(
            \mathbf z_i^{\mathrm{src}}
            \!\cdot\!
            \mathbf z_k^{\mathrm{dst}}
        \bigr)
    },
\end{equation}

\begin{equation}
    \widehat{\mathbf Q}^{\mathrm{in}}_{i,j}
    =
    \frac{
        \exp\!\bigl(
            \mathbf z_i^{\mathrm{dst}}
            \!\cdot\!
            \mathbf z_j^{\mathrm{src}}
        \bigr)
    }{
        \sum_{k=1}^{N}
        \exp\!\bigl(
            \mathbf z_i^{\mathrm{dst}}
            \!\cdot\!
            \mathbf z_k^{\mathrm{src}}
        \bigr)
    }.
    \label{eq:trip-pred}
\end{equation}

The predicted transition distributions are matched against the empirical OD distributions in both directions using the following loss:
\begin{equation}
    \mathcal L_{\mathrm{mob}}
    =
    -\sum_{i,j}
    \Bigl[
        \mathbf M_{i,j}
        \log
        \widehat{\mathbf Q}^{\mathrm{out}}_{i,j}
        +
        \mathbf M_{j,i}
        \log
        \widehat{\mathbf Q}^{\mathrm{in}}_{i,j}
    \Bigr].
    \label{eq:mob-loss}
\end{equation}

\subsection{Total Training Objective}
\label{sec:total-loss}

MoSS is trained end-to-end by minimizing
\begin{equation}
    \mathcal{L}_{\mathrm{total}}
    = \mathcal{L}_{\mathrm{mob}}
        + \lambda_{\mathrm{align}}\, \mathcal{L}_{\mathrm{align}}
        + \lambda_{\mathrm{orth}}\,  \mathcal{L}_{\mathrm{orth}},
    \label{eq:total-loss}
\end{equation}
where $\lambda_{\mathrm{align}}, \lambda_{\mathrm{orth}} \geq 0$ control the strength of cross-view shared alignment and dual orthogonality terms, respectively.

\section{Experiments}
\label{sec:experiments}

\subsection{Datasets and Tasks}
\label{sec:datasets}

To evaluate the effectiveness of the proposed model \textbf{MoSS}, experiments are conducted using real-world data from two U.S. cities: New York City (NYC) and Chicago (CHI). Table~\ref{tab:dataset_stats} summarizes the datasets.

The region division uses $180$ census tracts in NYC and $77$ community areas in CHI. For each region, we aggregate taxi-trip records by pickup and drop-off region to form an hourly OD tensor, which serves as the raw mobility input to MoSS. The mobility data is obtained from the NYC Taxi \& Limousine Commission~\cite{nycod} for NYC and from the City of Chicago data portal~\cite{chicagodp} for CHI. The embeddings of the region are evaluated in three downstream prediction tasks.

\begin{itemize}
    \item \textbf{Crime prediction.} Predict the annual crime count of each region from its embedding.
    \item \textbf{Income prediction.} Predict the median household income of each region.
    \item \textbf{Service-call prediction.} Predict the annual 311 service-call count for each region.
\end{itemize}

\begin{table}[t]
\centering
\caption{Dataset statistics and sources. Counts (regions, trips, and events)
are reported as totals. Income is reported as the average of per-region
median household income.}
\label{tab:dataset_stats}
\resizebox{0.95\columnwidth}{!}{
\small
\setlength{\tabcolsep}{4pt}
\begin{tabular}{lccp{6.0cm}}
\toprule
Data & NYC & CHI & Source \\
\midrule
Regions
& 180
& 77
& U.S.\ Census Bureau~\cite{uscensus} and CHIDP~\cite{chicagodp} \\

Taxi trips
& 9{,}779{,}714
& 3{,}368{,}049
& NYCOD~\cite{nycod} and CHIDP~\cite{chicagodp} \\

\midrule
Crime
& 35{,}335
& 18{,}200
& NYCOD~\cite{nycod} and CHIDP~\cite{chicagodp} \\

Income
& \$84{,}600
& \$74{,}734
& NYCOD~\cite{nycod} and Chicago Health Atlas~\cite{chicagohealthatlas} \\

Service calls
& 516{,}187
& 24{,}350
& NYCOD~\cite{nycod} and CHIDP~\cite{chicagodp} \\
\bottomrule
\end{tabular}
}
\end{table}

\subsection{Implementation Details}
\paragraph{Zigzag persistence.}
We extract only $H_0$ persistence, which tracks connectivity by counting connected components. For each region, we compute one zigzag $H_0$ diagram per direction (inflow/outflow), which yields two persistence-diagram views per region.

\paragraph{Persistence-diagram encoder.}
The PD encoder of Eq.~\eqref{eq:topo-encoder} takes each persistence point as a $3$-dim feature (birth, death, persistence) and applies $L_\phi = 4$ point-wise convolutions with widths $C_1 = 32$, $C_2 = 64$, $C_3 = 128$, and final width $D$, followed by max-pooling along the point axis.

\paragraph{Temporal encoder.}
The two outflow / inflow time series are stacked into a single 2-channel input and processed jointly by a dilated TCN with hidden width $C_h = 32$, $L_{\text{tcn}} = 10$ residual blocks with kernel size $k_{\text{ts}} = 3$, dilation $2^{\ell}$ at the $\ell$-th block, and max-pooling over the time axis to width $D$.

\paragraph{Optimization.}
Adam optimizer is used throughout. Hyperparameters are tuned separately for each city. For NYC, we set the learning rate to \(10^{-3}\), the alignment weight \(\lambda_{\text{align}}\) to \(1\), and the synergy rank \(R\) to \(8\). For Chicago, the learning rate is \(8 \times 10^{-4}\), \(\lambda_{\text{align}} = 50\), and \(R = 4\). Shared settings across both cities include \(\lambda_{\text{orth}} = 50\), output dimension \(H = 144\), view dimension \(D = 16\), and dropout rate \(0.1\).

\subsection{Baselines}
\label{sec:baselines}

We compare against six representative urban region embedding methods that span the main families used to integrate human mobility with auxiliary information.
\begin{itemize}
    \item \textbf{MVURE}~\cite{mvure} adds a self-attention layer over view-specific embeddings and fuses them with adaptive view weights.
    \item \textbf{MGFN}~\cite{mgfn} learns separate source and destination embeddings from a multi-graph fusion of typical-week OD patterns.
    \item \textbf{HREP}~\cite{hrep} couples a heterogeneous region encoder with prompt-based downstream specialization, supervised by mobility, geographical, and POI signals.
    \item \textbf{ReCP}~\cite{recp} contrasts attribute and mobility autoencoders with a dual prediction objective.
    \item \textbf{MVJC}~\cite{mvjc} adds a structure-aware contrastive objective to mitigate the false-negative problem among functionally similar regions.
    \item \textbf{ComSRE}~\cite{comsre} disentangles commonality and view-specific representations across multiple urban views through attention-based fusion and contrastive learning.
\end{itemize}

Table~\ref{tab:baseline_taxonomy} positions these baselines and MoSS along modality and fusion strategy.

\begin{table}[t]
\centering
\caption{Categorization of baselines and MoSS along modality and fusion strategy.}
\label{tab:baseline_taxonomy}
\resizebox{0.95\columnwidth}{!}{
\begin{tabular}{l l l}
\toprule
Method & Modality & Fusion strategy \\
\midrule
MVURE~\cite{mvure}       & mobility + POI + check-in            & attention-based \\
MGFN~\cite{mgfn}         & mobility                             & attention-based \\
HREP~\cite{hrep}         & mobility + POI + geographic neighbor & attention-based \\
ReCP~\cite{recp}         & mobility + POI                      & contrastive \\
MVJC~\cite{mvjc}         & mobility + POI + check-in            & contrastive \\
ComSRE~\cite{comsre}     & mobility + POI                      & shared--private decomposition \\
\midrule
\textbf{MoSS (ours)}     & \textbf{mobility}              & \textbf{shared--private + synergy} \\
\bottomrule
\end{tabular}
}
\end{table}

\subsection{Evaluation Protocol}
\label{sec:eval}

Following prior work~\cite{hafusion,hrep,recp,mvure}, we evaluate the frozen region embeddings with a Ridge regressor under 5-fold cross-validation, as the number of regions is small. We report the mean absolute error (MAE), root mean squared error (RMSE), and coefficient of determination $R^2$ as mean and standard deviation over 5 runs.

\subsection{Main Results}
\label{sec:main-results}

We compare \textsc{MoSS} against six recent region-representation baselines---MVURE, MGFN, HREP, ReCP, MVJC, and ComSRE---on three downstream prediction tasks (crime, income, service call) in two cities, New York City ($N{=}180$) and Chicago ($N{=}77$). All baselines are trained with their authors' released code. Each configuration is evaluated with five runs; we report mean~$\pm$~std of MAE, RMSE, and~$R^2$ in Table~\ref{tab:main}. 

\paragraph{Overall performance.}
\textsc{MoSS} achieves state-of-the-art performance on every task in terms of MAE, RMSE, and $R^2$. The largest $R^2$ gains over the strongest per-column baseline appear on service call ($+0.040$ on NY over MVURE; $+0.100$ on CHI over HREP) and on income on CHI ($+0.069$ over HREP/MGFN). On crime, RMSE drops by $12.0\%$ on NY (HREP, $88.43 \to 77.82$) and $4.4\%$ on CHI (ComSRE, $117.94 \to 112.77$). Crucially, the same model configuration and the same set of region embeddings are used for all three tasks within a city, so the uniform gains reflect a more \emph{transferable} representation rather than per-task tuning---several baselines that are competitive on one task collapse on another (e.g.\ ComSRE is the strongest baseline on CHI crime, $R^2{=}0.548$, but falls to $0.187$ on CHI income; ReCP's CHI service call $R^2$ varies by $\pm 0.157$).

\paragraph{Cross-city consistency.}
Existing baselines exhibit performance variability across cities and tasks. For instance, ComSRE achieves the best performance on CHI crime prediction ($R^2{=}0.548$) but drops to the middle of the pack on NY crime ($0.450$, compared to HREP's $0.642$). Likewise, MGFN attains strong performance on CHI income prediction ($R^2{=}0.630$) yet degrades markedly on NY income ($0.460$), while HREP, which leads on NY crime ($0.642$), is overtaken on CHI crime ($0.513$). In contrast, \textsc{MoSS} ranks first in mean across all tasks and both cities, demonstrating stable generalization across different urban structures, prediction targets, and target scales.

\paragraph{Stability.}
\textsc{MoSS} attains the best mean on every NY task together with consistently low seed variance --- the lowest on NY crime and service call --- and remains competitive in variance on CHI while maintaining state-of-the-art predictive performance. In comparison, several baselines show either larger variability or lower overall accuracy depending on the task. For example, ReCP on CHI service call exhibits a large fluctuation ($\pm 0.157$), while MGFN on CHI service call and MVURE across CHI tasks also display relatively high variance (0.205 and 0.087--0.094, respectively). Conversely, some methods such as ComSRE and MVJC achieve relatively small standard deviations on NY tasks (0.020), but with lower mean performance than \textsc{MoSS}.

\begin{table*}[t]
\centering
\caption{Downstream task performance on \textbf{(a) New York} and \textbf{(b) Chicago}. Each cell reports MAE\,$\downarrow$ /RMSE\,$\downarrow$ / $R^2$\,$\uparrow$ as mean\,$\pm$\,std over $5$ runs, evaluated with a frozen embedding and a Ridge regressor under 5-fold cross-validation; best results in \textbf{bold}.}
\label{tab:main}
\subcaption{New York City ($N = 180$ regions)}
\label{tab:main_manh}
\resizebox{\textwidth}{!}{%
\begin{tabular}{l ccc ccc ccc}
\toprule
& \multicolumn{3}{c}{Crime}
& \multicolumn{3}{c}{Income}
& \multicolumn{3}{c}{Service Call} \\
\cmidrule(lr){2-4}\cmidrule(lr){5-7}\cmidrule(lr){8-10}
Model
& MAE\,$\downarrow$ & RMSE\,$\downarrow$ & $R^2$\,$\uparrow$
& MAE\,$\downarrow$ & RMSE\,$\downarrow$ & $R^2$\,$\uparrow$
& MAE\,$\downarrow$ & RMSE\,$\downarrow$ & $R^2$\,$\uparrow$ \\
\midrule
MVURE
& {\small 66.97\,$\pm$\,2.53} & {\small 91.32\,$\pm$\,3.11} & {\small 0.618\,$\pm$\,0.026}
& {\small 24{,}679.25\,$\pm$\,555.00} & {\small 34{,}039.77\,$\pm$\,1{,}419.32} & {\small 0.454\,$\pm$\,0.046}
& {\small 1{,}388.52\,$\pm$\,73.98} & {\small 2{,}119.61\,$\pm$\,79.52} & {\small 0.402\,$\pm$\,0.045} \\
MGFN
& {\small 72.65\,$\pm$\,1.91} & {\small 95.92\,$\pm$\,3.33} & {\small 0.579\,$\pm$\,0.029}
& {\small 25{,}348.51\,$\pm$\,453.34} & {\small 33{,}892.86\,$\pm$\,906.87} & {\small 0.460\,$\pm$\,0.029}
& {\small 1{,}561.49\,$\pm$\,64.51} & {\small 2{,}297.65\,$\pm$\,74.43} & {\small 0.297\,$\pm$\,0.046} \\
HREP
& {\small 67.52\,$\pm$\,3.40} & {\small 88.43\,$\pm$\,3.82} & {\small 0.642\,$\pm$\,0.030}
& {\small 24{,}050.41\,$\pm$\,609.86} & {\small 32{,}811.26\,$\pm$\,613.64} & {\small 0.494\,$\pm$\,0.019}
& {\small 1{,}421.57\,$\pm$\,59.56} & {\small 2{,}174.02\,$\pm$\,53.01} & {\small 0.371\,$\pm$\,0.031} \\
ReCP
& {\small 78.47\,$\pm$\,2.25} & {\small 105.62\,$\pm$\,3.71} & {\small 0.489\,$\pm$\,0.035}
& {\small 23{,}591.51\,$\pm$\,786.42} & {\small 33{,}752.57\,$\pm$\,904.24} & {\small 0.464\,$\pm$\,0.029}
& {\small 1{,}638.69\,$\pm$\,14.16} & {\small 2{,}379.75\,$\pm$\,92.45} & {\small 0.246\,$\pm$\,0.058} \\
MVJC
& {\small 96.56\,$\pm$\,1.61} & {\small 126.20\,$\pm$\,1.45} & {\small 0.272\,$\pm$\,0.017}
& {\small 24{,}854.67\,$\pm$\,395.86} & {\small 34{,}048.97\,$\pm$\,623.97} & {\small 0.455\,$\pm$\,0.020}
& {\small 1{,}730.27\,$\pm$\,32.07} & {\small 2{,}550.76\,$\pm$\,19.69} & {\small 0.135\,$\pm$\,0.013} \\
ComSRE
& {\small 83.32\,$\pm$\,1.23} & {\small 109.62\,$\pm$\,1.67} & {\small 0.450\,$\pm$\,0.017}
& {\small 25{,}972.68\,$\pm$\,209.84} & {\small 36{,}785.39\,$\pm$\,352.58} & {\small 0.364\,$\pm$\,0.012}
& {\small 1{,}527.64\,$\pm$\,21.50} & {\small 2{,}233.19\,$\pm$\,27.18} & {\small 0.337\,$\pm$\,0.016} \\
\midrule
\textbf{MoSS (ours)}
& \textbf{58.54\,$\pm$\,1.47} & \textbf{77.82\,$\pm$\,1.88} & \textbf{0.723\,$\pm$\,0.013}
& \textbf{22{,}570.61\,$\pm$\,553.36} & \textbf{31{,}961.66\,$\pm$\,546.65} & \textbf{0.520\,$\pm$\,0.017}
& \textbf{1{,}350.51\,$\pm$\,25.85} & \textbf{2{,}048.49\,$\pm$\,17.70} & \textbf{0.442\,$\pm$\,0.010} \\
\bottomrule
\end{tabular}}
\vspace{1em}

\subcaption{Chicago ($N = 77$ community areas)}
\label{tab:main_chi}
\resizebox{\textwidth}{!}{%
\begin{tabular}{l ccc ccc ccc}
\toprule
& \multicolumn{3}{c}{Crime}
& \multicolumn{3}{c}{Income}
& \multicolumn{3}{c}{Service Call} \\
\cmidrule(lr){2-4}\cmidrule(lr){5-7}\cmidrule(lr){8-10}
Model
& MAE\,$\downarrow$ & RMSE\,$\downarrow$ & $R^2$\,$\uparrow$
& MAE\,$\downarrow$ & RMSE\,$\downarrow$ & $R^2$\,$\uparrow$
& MAE\,$\downarrow$ & RMSE\,$\downarrow$ & $R^2$\,$\uparrow$ \\
\midrule
MVURE
& {\small 106.65\,$\pm$\,9.33} & {\small 137.63\,$\pm$\,11.04} & {\small 0.386\,$\pm$\,0.094}
& {\small 17{,}596.48\,$\pm$\,1{,}338.95} & {\small 22{,}888.09\,$\pm$\,1{,}753.51} & {\small 0.436\,$\pm$\,0.087}
& {\small 196.32\,$\pm$\,15.70} & {\small 277.74\,$\pm$\,21.96} & {\small 0.397\,$\pm$\,0.093} \\
MGFN
& {\small 105.98\,$\pm$\,4.95} & {\small 141.99\,$\pm$\,7.58} & {\small 0.348\,$\pm$\,0.069}
& {\small 14{,}251.82\,$\pm$\,1{,}337.33} & {\small 18{,}500.46\,$\pm$\,1{,}630.94} & {\small 0.630\,$\pm$\,0.067}
& {\small 230.35\,$\pm$\,29.03} & {\small 327.00\,$\pm$\,38.85} & {\small 0.158\,$\pm$\,0.205} \\
HREP
& {\small 91.31\,$\pm$\,5.31} & {\small 122.80\,$\pm$\,5.09} & {\small 0.513\,$\pm$\,0.041}
& {\small 13{,}985.99\,$\pm$\,1{,}176.41} & {\small 18{,}506.64\,$\pm$\,1{,}676.33} & {\small 0.630\,$\pm$\,0.068}
& {\small 178.36\,$\pm$\,4.59} & {\small 253.41\,$\pm$\,8.26} & {\small 0.501\,$\pm$\,0.032} \\
ReCP
& {\small 99.22\,$\pm$\,8.82} & {\small 135.96\,$\pm$\,8.27} & {\small 0.402\,$\pm$\,0.073}
& {\small 16{,}038.75\,$\pm$\,1{,}486.26} & {\small 22{,}514.94\,$\pm$\,1{,}845.25} & {\small 0.453\,$\pm$\,0.088}
& {\small 195.54\,$\pm$\,26.83} & {\small 280.62\,$\pm$\,37.17} & {\small 0.378\,$\pm$\,0.157} \\
MVJC
& {\small 114.08\,$\pm$\,3.68} & {\small 148.21\,$\pm$\,3.22} & {\small 0.292\,$\pm$\,0.030}
& {\small 16{,}878.63\,$\pm$\,433.16} & {\small 21{,}639.37\,$\pm$\,528.62} & {\small 0.498\,$\pm$\,0.025}
& {\small 185.13\,$\pm$\,8.00} & {\small 271.29\,$\pm$\,10.93} & {\small 0.428\,$\pm$\,0.045} \\
ComSRE
& {\small 83.85\,$\pm$\,4.88} & {\small 117.94\,$\pm$\,10.37} & {\small 0.548\,$\pm$\,0.077}
& {\small 20{,}423.16\,$\pm$\,685.85} & {\small 27{,}524.31\,$\pm$\,1{,}337.32} & {\small 0.187\,$\pm$\,0.081}
& {\small 236.27\,$\pm$\,3.21} & {\small 314.85\,$\pm$\,5.53} & {\small 0.230\,$\pm$\,0.027} \\
\midrule
\textbf{MoSS (ours)}
& \textbf{82.45\,$\pm$\,7.90} & \textbf{112.77\,$\pm$\,9.88} & \textbf{0.587\,$\pm$\,0.072}
& \textbf{12{,}485.64\,$\pm$\,869.83} & \textbf{16{,}733.43\,$\pm$\,954.44} & \textbf{0.699\,$\pm$\,0.035}
& \textbf{156.69\,$\pm$\,5.88} & \textbf{226.57\,$\pm$\,10.48} & \textbf{0.601\,$\pm$\,0.037} \\

\bottomrule
\end{tabular}}
\vspace{0.5em}
\footnotesize
\end{table*}

\subsection{Ablation Study}
\label{sec:ablation}

To isolate the contribution of each component of our model, we ablate four
modules:
\begin{itemize}
    \item \textbf{w/o Seq.} Replace the dilated-TCN encoding of each region's hourly inflow/outflow \emph{time series} with a row-wise embedding of its row in the (row-normalized) OD matrix.
    \item \textbf{w/o Strct.} Remove structure stream.
    \item \textbf{w/o SP.} Remove the shared--private decomposition: the three raw view embeddings feed the synergy module directly, with no shared embedding and no alignment/orthogonality regularization.
    \item \textbf{w/o Syn.} Replace the synergy module with a plain concatenation of the shared embeddings and the three private embeddings.
\end{itemize}
The results for all four variants are shown in Table~\ref{tab:ablation} and we make the following observations.

\begin{table}[t]
\centering
\caption{Ablation study on NY and CHI. We remove or replace one module at a time: \textbf{w/o Seq} replaces the dilated-TCN sequence encoder with a row-wise OD-row embedding, \textbf{w/o Strct} drops the structure stream, \textbf{w/o SP} removes the shared--private decomposition (feeding raw view embeddings into the synergy module), and \textbf{w/o Syn} replaces the synergy module with plain concatenation. Each cell reports per-task $R^2$\,$\uparrow$ (mean over 5 runs); best results in \textbf{bold}.}
\label{tab:ablation}
\begin{tabular}{l ccc ccc}
\toprule
& \multicolumn{3}{c}{NY} & \multicolumn{3}{c}{CHI} \\
\cmidrule(lr){2-4} \cmidrule(lr){5-7}
Variant & Crime & Income & Service Call & Crime & Income & Service Call \\
\midrule
Full      & \textbf{0.723} & \textbf{0.520} & \textbf{0.442} & \textbf{0.587} & \textbf{0.699} & \textbf{0.601} \\
w/o Seq   & 0.558          & 0.441          & 0.378          & 0.404          & 0.577          & 0.284 \\
w/o Strct & 0.646          & 0.481          & 0.406          & 0.442          & 0.504          & 0.422 \\
w/o SP    & 0.622          & 0.432          & 0.395          & 0.208          & 0.576          & 0.326 \\
w/o Syn   & 0.582          & 0.477          & 0.395          & 0.414          & 0.454          & 0.426 \\
\bottomrule
\end{tabular}%
\end{table}

\paragraph{Effectiveness of sequence stream.}
Replacing the sequence stream with a row-wise OD-row embedding (\textbf{w/o Seq}) degrades the model, with magnitude varying by city and task. The resulting drop in downstream performance shows that the time-evolving volume semantics carries predictive information that neither the aggregate OD-row embedding nor the topological views recover.

\paragraph{Effectiveness of structure stream.}
Dropping the persistence-diagram views (\textbf{w/o Strct}) consistently degrades performance across all settings. Prediction across cities and tasks empirically confirms that this persistent connectivity structure carries signal that neither the raw temporal sequence nor a single-view embedding can substitute.

\paragraph{Effectiveness of shared--private decomposition.}
Removing the shared--private decomposition (\textbf{w/o SP}) feeds the raw view embeddings directly into the synergy module, with no shared embedding and no alignment/orthogonality regularization. The drop is consistent on NYC (e.g.\ Crime $0.723 \to 0.622$, Income $0.520 \to 0.432$) and particularly severe on CHI Crime ($0.587 \to 0.208$). This shows that the synergy module benefits from private embeddings that are disentangled from a common shared component, through shared--private decomposition.

\paragraph{Effectiveness of synergy module.}
Replacing the multilinear interaction with a plain concatenation of the shared embedding and the three private embeddings (\textbf{w/o Syn}) also degrades performance across both cities. This confirms that the performance gain does not arise merely from combining multiple view embeddings but from explicitly modeling their multiplicative interactions through the synergy mechanism.


\subsection{Hyperparameter Analysis}
\label{sec:hyper}

\begin{figure}[htbp]
\centering
\includegraphics{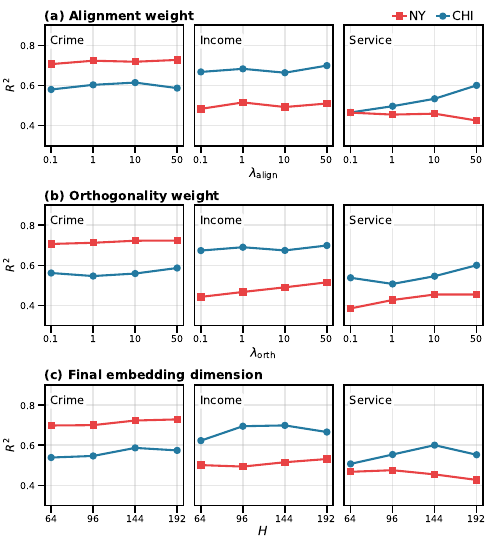}
\caption{Impact of (a) alignment weight $\lambda_{\text{align}}$, (b) orthogonality weight $\lambda_{\text{orth}}$, and (c) final embedding dimension $H$, on NYC and CHI. Each curve reports per-task $R^2$\,$\uparrow$.}
\label{fig:hyper_fig}
\end{figure}

We analyze \textsc{MoSS}'s sensitivity to the following hyperparameters.
\begin{itemize}
    \item \textbf{Alignment weight} $\lambda_{\text{align}} \in \{0.1,\ 1,\ 10,\ 50\}$.
    \item \textbf{Orthogonality weight} $\lambda_{\text{orth}} \in \{0.1,\ 1,\ 10,\ 50\}$.
    \item \textbf{Final embedding dimension} $H \in \{64,\ 96,\ 144,\ 192\}$.
\end{itemize}

\paragraph{Weight parameters $\lambda_{\text{align}}$ and $\lambda_{\text{orth}}$.}
The two weights affect the three tasks differently (Figure~\ref{fig:hyper_fig}a,b). Crime is the most robust, varying within $0.02$ in $R^2$ across both weight ranges on both cities. Service call is the most responsive, and the two cities respond in opposite directions to $\lambda_{\text{align}}$: CHI rises monotonically from $0.466$ at $\lambda_{\text{align}}{=}0.1$ to $0.601$ at $\lambda_{\text{align}}{=}50$, while NY declines from $0.465$ to $0.426$ across the same range.

\paragraph{Final embedding dimension.}
On CHI, all three tasks peak at $H=144$ and decline at $H=192$, with the worst performance at $H=64$ (Figure~\ref{fig:hyper_fig}c). NY shows two different patterns: crime and income improve monotonically up to $H=192$ ($0.699 \to 0.729$ and $0.501 \to 0.532$), whereas service call peaks early at $H=96$ ($0.476$) and declines at larger dimensions.

\subsection{Model Size and Training Efficiency}
\label{sec:efficiency}

We additionally compare \textsc{MoSS} against the six baselines along two practical axes: number of trainable parameters and wall-clock time per
training epoch. Parameter counts are obtained by enumerating all trainable tensors in each model after instantiation with the hyperparameters used to produce Table~\ref{tab:main}. Per-epoch time is measured on a single NVIDIA RTX~3090 (24~GB) with $20$ measured epochs after $3$ warm-up epochs. Numbers are reported in Table~\ref{tab:cost}.

\begin{table}[t]
\centering
\caption{Trainable parameters and per-epoch training time of \textsc{MoSS} and the six baselines on NY and CHI; lower is better for both metrics.}
\label{tab:cost}
\begin{tabular}{l r r r r}
\toprule
& \multicolumn{2}{c}{NY ($N=180$)} & \multicolumn{2}{c}{CHI ($N=77$)} \\
\cmidrule(lr){2-3}\cmidrule(lr){4-5}
Model & Params & ms/epoch & Params & ms/epoch \\
\midrule
MVURE     &    222\,K &  91.0 &    222\,K &  23.7 \\
MGFN      &  13.39\,M &  27.3 &   3.46\,M &  24.5 \\
HREP      &    188\,K &  26.6 &    188\,K &  26.2 \\
ReCP      &    289\,K & 155.4 &    235\,K & 166.9 \\
MVJC      &    829\,K &  51.4 &    558\,K &  49.5 \\
ComSRE    &    458\,K &  29.4 &    353\,K &  28.7 \\
\midrule
\textbf{MoSS (ours)} & 157\,K & 62.0 & 157\,K & 61.6 \\
\bottomrule
\end{tabular}
\end{table}

\paragraph{Parameter count.}
\textsc{MoSS} uses $\sim$157K trainable parameters on both cities, the smallest among the compared methods. It uses fewer parameters than HREP ($188$K), and is $85\times$ smaller than MGFN on NY ($13.39$M $\to$ $157$K) and $22\times$ smaller on CHI ($3.46$M $\to$ $157$K).

\paragraph{Training time.}
\textsc{MoSS} incurs a moderately higher per-epoch cost than most baselines, with only MVURE (NY) and ReCP being slower, because its sequence and structure encoders together with the multi-degree synergy fusion add per-step overhead beyond plain attention or contrastive objectives. This reflects a deliberate trade-off: these components raise the per-epoch cost yet keep \textsc{MoSS} compact in parameters and drive its accuracy gains.

\section{Conclusion}
\label{sec:conclusion}

We presented MoSS, an urban region embedding framework that combines mobility time series with zigzag persistence diagrams of the time-evolving connectivity-graphs. A shared--private decomposition, paired with cross-view alignment and orthogonality losses, fuses these complementary signals into a unified region representation refined by multi-degree interaction. Although MoSS draws solely on mobility data, it outperforms urban region embedding baselines that additionally rely on auxiliary modalities such as POI and check-in, while using a fraction of the parameters of the strongest baselines. We view the temporal--topological pairing as a general design pattern for representing urban regions and leave its extension to additional modalities, longer time scales, and cross-city transfer for future work.




\bibliographystyle{IEEEtran}
\bibliography{IEEEabrv,Bibliography}

\end{document}